\documentclass{article}

\PassOptionsToPackage{numbers,sort&compress}{natbib}
\usepackage[dblblindworkshop,final]{neurips_2026}

\usepackage[utf8]{inputenc}
\usepackage[T1]{fontenc}
\usepackage{hyperref}
\usepackage{url}
\usepackage{booktabs}
\usepackage{amsfonts}
\usepackage{amsmath}
\usepackage{algorithm}
\usepackage{algpseudocode}
\usepackage{microtype}
\usepackage{xcolor}
\usepackage{graphicx}

\title{ULPack: Reference-Conditioned Lossless Compression of Checkpoint Trajectories}
\workshoptitle{Neural Network Artifacts as a New Data Modality}

\author{%
  Aniket Srinivasan Ashok\\
  University of Waterloo\\
  \texttt{a3sriniv@uwaterloo.ca}
}

\begin{document}

\maketitle

\begin{abstract}
Studying model training requires checkpoint trajectories, whose storage cost
limits temporal resolution. Although adjacent checkpoints are numerically close,
existing lossless codecs either operate on XOR bit patterns or ignore
trajectory-specific conditional structure. We show that a reference weight's
exponent predicts the bit length of its displacement in ordered floating-point
space without side information. We introduce \textbf{ULPack}, a lossless codec which exploits this dependence by
conditioning entropy models of ordered-integer residuals on the reference exponent.
Across bfloat16, float16, and float32 trajectories from eight model families,
ULPack is the smallest codec on every evaluated checkpoint pair.
Averaged over each pre-training sequence, it reduces storage by up to 14.9\%
against ZipNN, 15.0\% against zstd-19, and 9.2\% against FM-Delta; in RL
fine-tuning it is 29.8\% smaller than the best baseline.
Compression improves as updates shrink relative to the weights and
reference conditioning strengthens, while near-uniform low-order bits set the
remaining compression floor and earlier checkpoints add little information.
\end{abstract}

\section{Introduction}

Training runs are increasingly released as sequences of checkpoints rather than
as single final models. Pythia, MultiBERTs, Amber, and OLMo each publish tens to
hundreds of intermediate checkpoints
\citep{biderman2023pythia,sellam2022multiberts,liu2023llm360,groeneveld2024olmo},
because questions about when a capability appears, how a representation forms,
or whether a run reproduces concern the trajectory rather than its endpoint.
Retaining those trajectories quickly becomes expensive: a BLOOM-176B checkpoint
is 329\,GiB of bfloat16 weights \citep{lescao2022bloom}, so keeping one every
thousand steps of its 93{,}000-step run would require roughly 30\,TiB. At
population scale, storage limits both the temporal resolution of each run and
the number of trajectories that can be preserved.

Consecutive checkpoints are numerically close, making delta compression a
natural solution. Many lossless checkpoint compressors operate in the bit
domain, XORing corresponding floating-point representations before entropy
coding \citep{waddington2025lmc,hershcovitch2024zipnn,pelkonen2015gorilla}.
But bit disagreement need not reflect numerical distance: in float16,
\texttt{0x3BFF} and \texttt{0x3C00} are adjacent values, yet their XOR sets
eleven of sixteen bits. Ordered-integer differencing avoids this mismatch by
mapping floating-point values monotonically to integers before taking deltas,
an established approach in floating-point and model compression
\citep{lindstrom2006fpzip,ning2024fm}.

Our question is what additional structure appears when these model pairs are
successive states of an \emph{optimization trajectory}. We find that the scale
of a weight's rank-space displacement is strongly predicted by the exponent of
the corresponding reference weight, and that this dependence strengthens as
training progresses. Over 2.47\,B elements of a complete BLOOM
90k$\rightarrow$91k shard delta, the reference exponent explains 20.2\% of
residual bit-length entropy, compared with 0.003\% for a spatially adjacent
element. Because the decoder already holds the reference checkpoint, this
context requires no per-element side information. ULPack exploits this
trajectory-specific conditional structure rather than treating ordered-integer
residuals as an unconditional stream.

This paper contributes:

\begin{itemize}

\item We identify and quantify trajectory-specific dependence between the
reference exponent and residual scale, while finding little additional
information in spatial context or longer temporal history.

\item We introduce \textbf{ULPack}, a lossless codec that exploits this
dependence through exponent-conditioned coding of ordered-integer residuals,
improving over ZipNN, zstd-19, and FM-Delta by up to 14.9\%, 15.0\%, and 9.2\%
on average in pre-training and by 29.8\% over the best baseline in RL fine-tuning.

\item We characterize how this structure evolves through training, across
model scales from 70M to 12B parameters, and in RL fine-tuning: compression
and reference predictiveness increase as updates shrink, while near-uniform
raw-tail bits increasingly determine encoded size.

\end{itemize}

\section{Background and related work}

\paragraph{Lossless model and checkpoint compression.}
LMC \citep{waddington2025lmc} compresses consecutive checkpoints using
byte-grouped XOR deltas followed by run-length and blockwise Huffman coding,
while ZipNN \citep{hershcovitch2024zipnn} compresses individual checkpoints by
separating floating-point bytes according to significance. XOR differencing is
also standard in floating-point time-series compression
\citep{pelkonen2015gorilla}. FM-Delta \citep{ning2024fm} instead maps
floating-point parameters to monotonic integers, differences each fine-tuned
model from a shared pretrained reference, entropy-codes the sign and
most-significant-bit position of each integer delta, and writes the remaining
low bits raw. Brevis \citep{shi2026brevis} takes a complementary approach,
synthesizing reversible transforms for self-contained tensors.

\paragraph{Floating-point residuals and training trajectories.}
Order-preserving float-to-integer transforms and residual coding predate neural
checkpoint compression. fpzip \citep{lindstrom2006fpzip} maps floating-point
values to monotonic integers and codes residuals against a spatial predictor;
VIREL \citep{zhang2026virel} similarly operates in an integer residual domain
for structured time-series data. Two-part codes that model a predictable
prefix while emitting less structured low-order bits are likewise established
\citep{weinberger2000locoi}. Our focus is therefore not the ordered-integer
mapping itself, but the conditional structure specific to optimization
trajectories: we measure how the co-located reference value predicts the next
rank-space displacement, how that structure evolves through training, and how
much additional information remains in longer temporal history. Checkpoint
suites such as Pythia and MultiBERTs
\citep{biderman2023pythia,sellam2022multiberts} and population-level Model
Zoos \citep{schurholt2022modelzoos} motivate treating such intermediate model
states as data rather than disposable training artifacts.

\paragraph{Lossy checkpoint compression.}
Quantization and pruning achieve higher ratios
\citep{chen2020lccheckpoint,eisenman2022checknrun,li2024excp} but do not
reproduce the trajectory exactly; Kim and Belyaev \citep{kim2025prediction}
also condition on the previous checkpoint's co-located weight, but code pruned
and quantized symbols.

\section{Method}
\label{sec:method}

A checkpoint sequence is $X_1,\dots,X_T$. We encode each tensor in $X_t$
against its corresponding tensor in $X_{t-1}$, the \emph{reference}; $X_1$ is
stored independently.

\subsection{Ordered-integer residuals}

Following prior floating-point and model-delta compressors
\citep{lindstrom2006fpzip,ning2024fm}, we first map floating-point bit patterns
to monotonic integer keys before differencing them. Let $\varphi$ map a $w$-bit
IEEE 754 representation bijectively to an unsigned integer key by flipping the
sign bit for nonnegative values and complementing all bits for negative ones.
For target value $x$ and reference value $r$, ULPack forms

\[
  \delta =
  \operatorname{signed}_w\!\left(
    \varphi(x)-\varphi(r) \bmod 2^w
  \right),
  \qquad
  z=\operatorname{zigzag}(\delta),
\]

where $\operatorname{signed}_w$ interprets a $w$-bit word as a two's-complement
integer, and ZigZag maps signed integers to nonnegative integers in increasing
magnitude ($0\mapsto0$, $-1\mapsto1$, $1\mapsto2$, \ldots), preserving numerical locality. We call $z$ the
\emph{ULP residual}.

The transform is exactly reversible. ULPack's modular formulation, inverse
mapping, and behavior on signed zeros, infinities, and NaN payloads are detailed
in Appendix~\ref{app:ulp-details}.

\subsection{Exponent-conditioned residual coding}

\textbf{The residual coding. } Ordered-integer residuals expose numerical locality, but their magnitudes still
vary substantially. ULPack separates the predictable high-order structure of a
residual from its increasingly uniform low-order bits.

For $z>0$, let $b=\operatorname{bitlen}(z)$ and write
\[
z =
\underbrace{1}_{\text{implicit}}
\ \ \underbrace{q_1\,q_2}_{\text{modeled}}
\ \ \underbrace{r_1\cdots r_{b-3}}_{\text{raw tail}},
\]
with absent fields omitted for short residuals. ULPack codes $b$ and up to two
bits below the leading one using rANS, while the remaining low-order bits
$r_i$ are written verbatim.

\begin{figure*}[t]
\centering
\includegraphics[width=\textwidth]{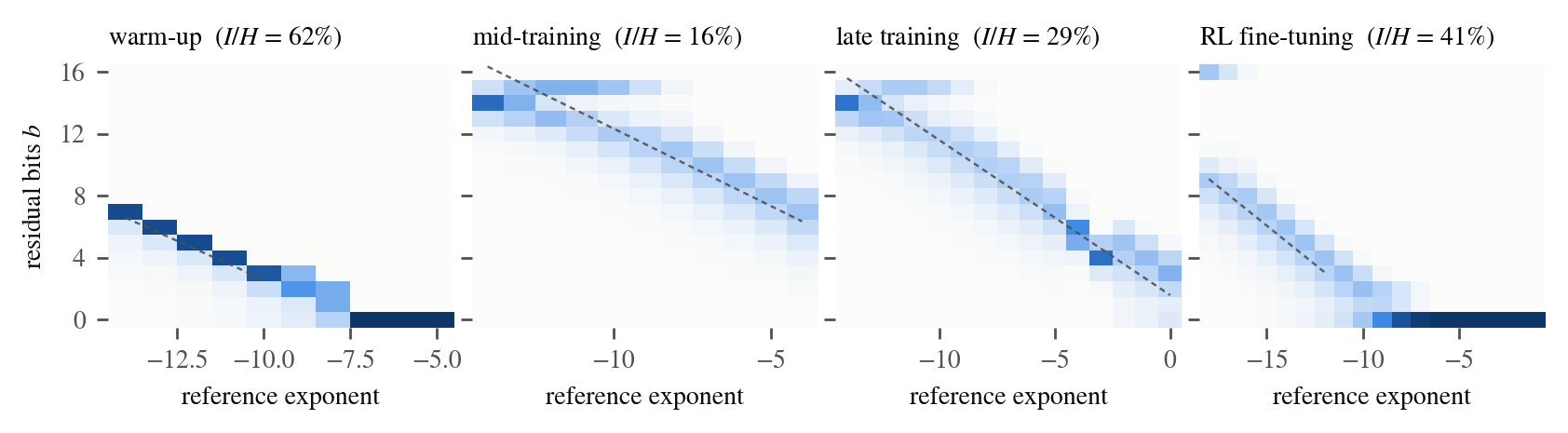}
\caption{Residual bit length conditioned on the exponent of the co-located
reference weight in warm-up, mid-training, late training, and RL fine-tuning.
The ridge shifts systematically with reference magnitude; when updates are
small, large weights do not change while small weights do.}
\label{fig:context}
\end{figure*}

Let $e_r$ denote the exponent of the corresponding reference value. The coded
symbols use the conditional models
\[
P(b\mid e_r), \qquad
P(q_1\mid e_r,b), \qquad
P(q_2\mid e_r,b,q_1).
\]
Because the decoder already holds the reference checkpoint, $e_r$ requires no
per-element side information. Figure~\ref{fig:context} shows that it strongly
stratifies residual scale. Probability tables are estimated per tensor and encoded with static rANS;
table construction and stream layout are detailed in
Appendix~\ref{app:coding}.

\textbf{Residual exponent bit-length entropy information. } We quantify the explained share as empirical normalized mutual information $\frac{I(b;e_r)}{H(b)}$. On the BLOOM 90k$\rightarrow$91k delta, $H(b)=3.0649$ and $I(b;e_r)=0.6183$ bits per weight, giving 20.2\%.

\section{Experimental setup}

\paragraph{Data.}
Table~\ref{tab:data} summarizes the evaluated trajectories; BLOOM, Amber, and
MultiBERTs follow datasets evaluated by LMC \citep{waddington2025lmc}. Exact
checkpoint pairs and identifiers are given in Appendix~\ref{app:experiments}.

\begin{table}[t]
\centering
\small
\caption{Checkpoint trajectories used for evaluation.}
\label{tab:data}
\begin{tabular}{lllrll}
\toprule
Sequence & Dtype & Size & Pairs & Evaluated unit & Payload \\
\midrule
BLOOM & bfloat16 & 176B & 19 & shard 36 & 4.93\,GB \\
Pythia & float16 & 70M--12B & 179 & complete models & 140.9\,MB--23.69\,GB \\
GPT-2 small (5 seeds) & float16 & 124M & 12 & complete models & 248.9\,MB \\
Amber & bfloat16 & 7B & 3 & complete models & 13.48\,GB \\
OLMoE & bfloat16 & 1B--7B & 3 & complete model & 13.84\,GB \\
OLMo & float32 & 7B & 5 & complete model & 27.55\,GB \\
MultiBERTs & float32 & 110M & 1 & complete model & 534.3\,MB \\
OLMo-2 RLVR & bfloat16 & 1B & 3 & complete model & 2.97\,GB \\
\bottomrule
\end{tabular}
\end{table}

\paragraph{Compute.}
Full-shard and ablation runs used a Linux VM with 30 Intel Xeon Platinum 8358
vCPUs and 222\,GiB RAM. Measurements were CPU-only and single-threaded unless
stated otherwise.

\paragraph{Baselines and ablations.}
We compare against ZipNN \citep{hershcovitch2024zipnn}, zstd-19, and FM-Delta
\citep{ning2024fm}. ZipNN and zstd operate on XOR residuals; FM-Delta is the
closest ordered-integer baseline. All methods are evaluated on the same
checkpoint pairs.

We use two controlled ablations. To isolate the residual representation, we
compress byte-grouped ULP residuals with the same zstd-19 backend used for XOR.
To isolate reference-exponent conditioning, we remove $e_r$ from ULPack's
probability models while holding its residual representation, decomposition,
and rANS coder fixed. Additional general-purpose compressors are reported in
Appendix~\ref{app:experiments}.

\paragraph{Metrics and verification.}
We report compressed delta size divided by uncompressed target payload; lower
is better. The independently stored anchor checkpoint is excluded. Relative
improvement over a baseline is $1-S_{\mathrm{ULPack}}/S_{\mathrm{base}}$.
Every output is reconstructed against its reference and verified byte-for-byte.
Runtime and implementation details are given in Appendix~\ref{app:experiments}.

\section{Results}
\label{sec:results}

\subsection{End-to-end compression}

Table~\ref{tab:main} summarizes end-to-end compression as the mean ratio over
all evaluated checkpoint pairs of each sequence. ULPack achieves the lowest
ratio on every sequence, improving over the strongest baseline by 2.2--9.2\% in
pre-training and 29.8\% in RL fine-tuning.

\begin{table}[t]
  \centering
  \footnotesize
  \setlength{\tabcolsep}{3.5pt}
  \caption{Mean compression ratio over checkpoint pairs (compressed size over
  payload; lower is better) for the pairs in Table~\ref{tab:data}. Pythia
  excludes the warm-up checkpoints and Pythia-1B (Appendix~\ref{app:coding}).
  Best in bold, second best underlined. Margin is ULPack's improvement over
  the strongest baseline.}
  \label{tab:main}
  \begin{tabular}{llrrrrr}
    \toprule
    Sequence & Format & \textbf{ULPack} & FM-Delta & ZipNN & XOR + zstd-19 & Margin \\
    \midrule
    \multicolumn{7}{l}{\textit{Pre-training}} \\
    BLOOM-176B & bfloat16 & \textbf{0.5134} & \underline{0.5553} & 0.5793 & 0.5756 & 7.5\% \\
    Pythia 70M--12B & float16 & \textbf{0.6444} & \underline{0.6822} & 0.7099 & 0.7108 & 5.5\% \\
    GPT-2 small & float16 & \textbf{0.6764} & \underline{0.7106} & 0.7311 & 0.7318 & 4.8\% \\
    Amber-7B & bfloat16 & \textbf{0.4672} & \underline{0.5144} & 0.5488 & 0.5497 & 9.2\% \\
    OLMoE-1B-7B & bfloat16 & \textbf{0.5810} & 0.6232 & \underline{0.6213} & 0.6222 & 6.5\% \\
    OLMo-7B & float32 & \textbf{0.7096} & \underline{0.7265} & 0.7535 & 0.7462 & 2.3\% \\
    MultiBERTs & float32 & \textbf{0.7920} & 0.8116 & \underline{0.8099} & 0.8172 & 2.2\% \\
    \midrule
    \multicolumn{7}{l}{\textit{RL fine-tuning}} \\
    OLMo-2 1B RLVR & bfloat16 & \textbf{0.0446} & \underline{0.0636} & 0.1607 & 0.0851 & 29.8\% \\
    \bottomrule
  \end{tabular}
\end{table}

\paragraph{Throughput.}
On one CPU core, ULPack is about as fast as FM-Delta but slower than ZipNN. An optimized
GPU implementation produces byte-identical output and, on one NVIDIA A10,
encodes at 0.5--1.0\,GB/s and decodes at 0.3--0.55\,GB/s, faster than
single-thread ZipNN at encoding on three of four models but slower at
decoding (Tables~\ref{tab:throughput} and~\ref{tab:throughput-new}).

\subsection{Where the gain comes from}
\label{sec:gain}

On the complete 3.50\,GB Amber shard, replacing XOR with ULP residuals under
the same zstd-19 backend reduces output by 6.9\%, and adding reference-exponent
conditioning to ULPack reduces its output by 9.7\%.
Table~\ref{tab:conditioning} generalizes the latter and reports the remaining
raw-tail floor.

\begin{table}[t]
\centering
\small
\caption{Where ULPack's gain comes from and where it stops. Conditioning gain
removes only the reference exponent; raw tail is emitted verbatim.}
\label{tab:conditioning}
\begin{tabular}{lrr}
\toprule
Sequence & Conditioning gain & Raw tail\\
\midrule
BLOOM (1k) & 11.1\% & 24.4\%\\
Amber & 9.7\% & 36.0\%\\
Pythia-70m & 6.1\% & 52.8\%\\
Pythia-6.9B & 6.8\% & 48.2\%\\
Pythia-12B & 6.7\% & 48.3\%\\
OLMo-7B & 2.4\% & 79.7\%\\
MultiBERTs & 1.6\% & 82.2\%\\
\bottomrule
\end{tabular}
\end{table}

Full ratios and bit-position diagnostics are reported in
Appendix~\ref{app:bit-domain}.

\subsection{How trajectory structure changes}
\label{sec:trajectory}

Two factors vary naturally along a checkpoint trajectory: the stage of
training and the distance between successive checkpoints. We study them
separately on complete BLOOM shard 36 and on complete Pythia checkpoints.

\begin{figure}[t]
  \centering
  \includegraphics[width=0.9\linewidth]{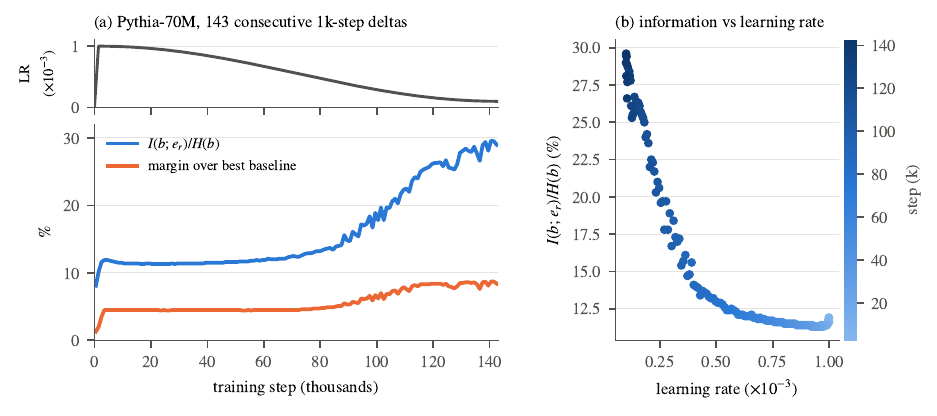}
  \caption{Pythia-70M across all 143 consecutive 1k-step deltas. Top: learning
  rate. Exponent information and ULPack's margin rise as the learning rate
  decays.}
  \label{fig:convergence}
\end{figure}

\paragraph{Across training.}
Later checkpoints are both more compressible and more predictable from the
reference. Over all 143 consecutive 1k-step deltas of Pythia-70M
(Figure~\ref{fig:convergence}), the ratio
falls from 0.835 to 0.516, the margin over the best baseline grows from 1.3\%
to 8.4\%, and $\frac{I(b;e_r)}{H(b)}$ grows from 8.1\% to 29.0\%. Both remain
nearly constant until about step 70k and rise as the learning rate decays.

\begin{figure*}[t]
\centering
\includegraphics[width=0.9\textwidth]{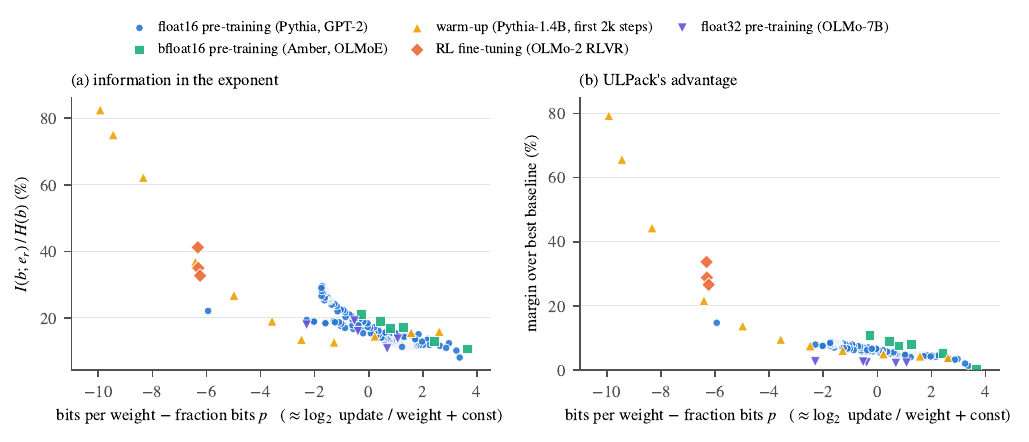}
\caption{Every completed checkpoint pair, by update size relative to the
weights. (a) Share of residual-length entropy explained by the reference
exponent. (b) ULPack's margin over the best baseline.}
\label{fig:regimes}
\end{figure*}

\paragraph{Across regimes.}
Figure~\ref{fig:regimes} places all completed pairs on one axis: ULPack's bits
per weight minus the format's fraction bits, which approximates the logarithm
of update size relative to weight size. Exponent information and ULPack's
margin both fall along this axis. Warm-up and RL fine-tuning, where updates
are smallest, reach margins of 27--79\%. Float32 carries similar exponent
information to 16-bit formats at the same position but gains less, because
most of its output is raw tail.

\paragraph{Checkpoint spacing.}
At eight matched endpoints, halving the BLOOM checkpoint gap from 10k to 5k
lowers ULPack's mean ratio from 0.5705 to 0.5267 (7.7\%). On Pythia-1.4B at
seven endpoints, each doubling of spacing from 1k to 10k adds 0.42--0.55 bits
per weight. Per-endpoint results are reported in
Appendix~\ref{app:trajectory-details}.

\subsection{Where compression stops helping}
\label{sec:limits}

\paragraph{The raw-tail ceiling.}
Across datasets, raw-tail share rises from 24.4\% to 82.2\% as conditioning
gain falls from 11.1\% to 1.6\% (Table~\ref{tab:conditioning}). Reference
conditioning increasingly predicts residual scale, but near-uniform low-order
bits leave progressively less structure to exploit. Further bit-budget and
per-tensor details are reported in Appendix~\ref{app:trajectory-details}.

\paragraph{Longer history adds little.}
Finally, we ask whether checkpoints earlier than the immediate reference
provide useful additional context. On BLOOM, the previous rank-space residual
explains 9.7\% of current residual-length entropy on its own, but only 0.02\%
after conditioning on the reference exponent. Successive raw rank-space
displacements are also only weakly anticorrelated ($\rho=-0.31$). Under a
Gaussian dependence model, this corresponds to 0.073 bits per weight of mutual
information, less than 1\% of a bfloat16 chain.

Consistent with this limited headroom, both bidirectional and momentum-based
reference schemes increase total stored size. Phase portraits
(Figure~\ref{fig:phase}) and detailed alternative-reference experiments are given in
Appendix~\ref{app:alternatives}.

\section{Discussion and limitations}

\paragraph{Evaluation scope.}
BLOOM is evaluated on a single shard rather than complete checkpoints. Amber,
OLMoE, and the RL run are each evaluated on only three checkpoint pairs.
ULPack's advantage is smallest at large learning rates and in float32. Where
almost no weight changes between checkpoints, or tensors are very small, its
per-tensor tables cost more than XOR+zstd-19 or FM-Delta.

\paragraph{Beyond weights.}
We evaluate model weights only; optimizer state may have substantially
different temporal structure. The fine-tuning experiments cover one GPT-2
pair and one RL run, while LoRA, non-transformer architectures, and float8
remain untested. Float8 is particularly interesting because its shorter
significand should reduce raw-tail dominance.

\section{Conclusion}

Ordered-integer residuals make numerical locality explicit; we show that
training trajectories contain additional structure within those residuals.
In particular, the exponent of the corresponding reference weight predicts
residual scale without per-element side information. ULPack exploits this
structure to outperform FM-Delta and bit-domain baselines across eight
public model families. Our trajectory analysis further shows that the remaining
cost increasingly lies in near-uniform low-order residual bits, while the
tested previous-step signal adds little beyond the immediate reference.

\newpage

\begin{ack}
    This project received no external funding or support, and the author declares no competing interests. 
\end{ack}

\bibliographystyle{abbrvnat}
\bibliography{refs}

\newpage
\appendix

\section{ULP residual details}
\label{app:ulp-details}

\paragraph{Order-preserving key.}
For a $w$-bit IEEE 754 representation $u$, let $h=2^{w-1}$. ULPack maps the
bit pattern to an unsigned integer key
\[
  \varphi(u)=
  \begin{cases}
    u \oplus h, & \text{if the sign bit is clear},\\
    \neg u, & \text{if the sign bit is set}.
  \end{cases}
\]
Positive bit patterns are already increasing as unsigned integers and are
lifted above the negative half of the key space. Negative bit patterns run in
the opposite direction and are therefore complemented. The inverse is
\[
  \varphi^{-1}(k)=
  \begin{cases}
    k\oplus h, & k\ge h,\\
    \neg k, & k<h.
  \end{cases}
\]
Thus $\varphi$ is a bijection over all $w$-bit patterns and agrees with
numerical order for non-NaN values. NaN payloads have no useful numerical
ordering, but still participate in the same exact bit-pattern bijection.

\paragraph{Modular displacement.}
For target value $x$ and reference value $r$, encoding computes
\[
  d=(\varphi(u_x)-\varphi(u_r))\bmod 2^w
\]
and interprets $d$ as a signed two's-complement integer $\delta$, so
\[
  \delta\in[-2^{w-1},2^{w-1}-1].
\]
Whenever the true rank displacement has magnitude below $2^{w-1}$,
$|\delta|$ is exactly the number of representable values traversed. Larger
displacements wrap around the key space; exact reconstruction is preserved
because decoding performs the corresponding modular addition.

The signed displacement is ZigZag-folded as
\[
  z =
  \begin{cases}
    2\delta, & \delta\ge 0,\\
    -2\delta-1, & \delta<0,
  \end{cases}
\]
so $-1$ and $+1$ map to the two smallest nonzero symbols, $1$ and $2$.

\paragraph{Residual transform pseudocode.}
Arithmetic below is performed in an unsigned $w$-bit type, so subtraction and
addition are modular.
\begin{algorithm}[h]
\caption{Bit-exact residual transform}
\label{alg:residual-transform}
\begin{algorithmic}[1]
\Procedure{EncodeResidual}{$x,r$}
    \State $k_x\gets\varphi(\operatorname{bits}(x))$
    \State $k_r\gets\varphi(\operatorname{bits}(r))$
    \State $\delta\gets\operatorname{signed}_w(k_x-k_r)$
    \State \Return $2\delta$ if $\delta\ge0$; otherwise $-2\delta-1$
\EndProcedure
\Procedure{DecodeResidual}{$z,r$}
    \State $\delta\gets z/2$ if $z$ is even; otherwise $-(z+1)/2$
    \State $k_x\gets\varphi(\operatorname{bits}(r))+\delta\pmod{2^w}$
    \State \Return $\operatorname{frombits}(\varphi^{-1}(k_x))$
\EndProcedure
\end{algorithmic}
\end{algorithm}

\paragraph{Losslessness and edge cases.}
For a fixed reference, $\varphi$, modular subtraction, and ZigZag folding are
all bijective, and the pseudocode applies their inverses in reverse order.
ULPack therefore round-trips every input bit pattern exactly, including signed
zeros, subnormals, infinities, and NaN payloads. No numerical tolerance or
floating-point arithmetic is involved.

\begin{figure}[h]
\centering
\includegraphics[width=0.75\linewidth]{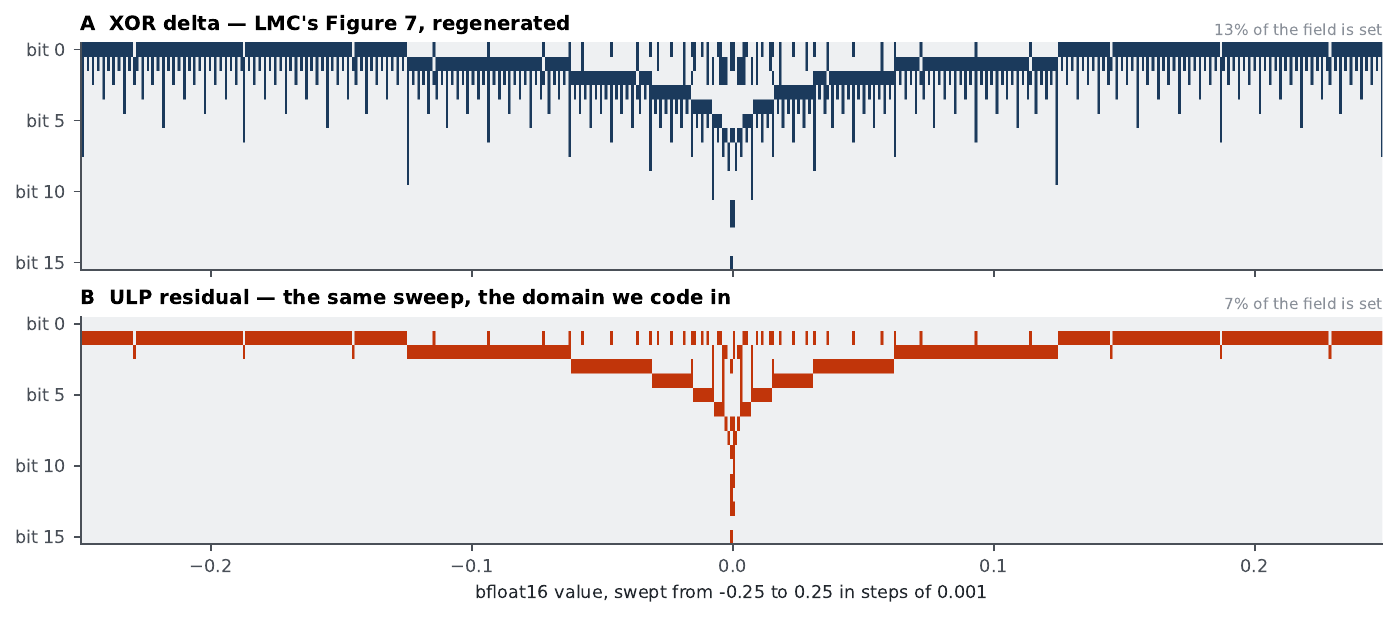}
\caption{A bfloat16 sweep under XOR (top) and ordered-integer differencing
(bottom). XOR produces irregular multi-bit changes even between adjacent
values, whereas ULP residuals preserve locality across exponent boundaries.
Marked-cell fraction falls from 13.5\% to 7.1\%.}
\label{fig:sweep}
\end{figure}

\section{Coding details}
\label{app:coding}

\paragraph{Symbols and contexts.}
For each ZigZag-folded residual $z_i$, let
\[
  b_i =
  \begin{cases}
    0, & z_i=0,\\
    \lfloor\log_2 z_i\rfloor+1, & z_i>0,
  \end{cases}
\]
and let $e_i$ be the exponent field of the corresponding reference value.
The leading one of a nonzero residual is implicit. ULPack models the next two
bits when present and writes all lower bits verbatim:
\[
\begin{array}{lll}
\text{length} & b_i & \text{context } e_i,\\
\text{refinement 1} & z_i[b_i-2] & \text{context } (e_i,b_i),\quad b_i\ge2,\\
\text{refinement 2} & z_i[b_i-3] & \text{context } (e_i,b_i,z_i[b_i-2]),
  \quad b_i\ge3,\\
\text{raw tail} & z_i[b_i-4:0] & \max(0,b_i-3)\text{ unmodeled bits}.
\end{array}
\]
Here $z[j]$ denotes bit $j$, with bit zero least significant. The decoder
already holds the reference tensor, so every $e_i$ is available without side
information. For float16, bfloat16, and float32, the length alphabets contain
17, 17, and 33 symbols respectively.

\paragraph{Tensor encoder pseudocode.}
\begin{algorithm}[h]
\caption{ULPack tensor encoder}
\label{alg:tensor-encoder}
\begin{algorithmic}[1]
\Procedure{EncodeTensor}{$X,R$}
    \For{each element $i$}
        \State $z_i\gets\Call{EncodeResidual}{X_i,R_i}$
        \State $e_i\gets\operatorname{exponent}(\operatorname{bits}(R_i))$
        \State $b_i\gets\operatorname{bitlength}(z_i)$
    \EndFor
    \State Estimate and transmit normalized tables for all modeled stages
    \State rANS-encode $b_i$ under context $e_i$
    \State rANS-encode refinement 1 under $(e_i,b_i)$ where $b_i\ge2$
    \State rANS-encode refinement 2 under $(e_i,b_i,\text{refinement 1})$
        where $b_i\ge3$
    \State Pack the lowest $\max(0,b_i-3)$ bits of each $z_i$ in element order
    \State \Return header, three coded stages, and packed raw tail
\EndProcedure
\end{algorithmic}
\end{algorithm}
Decoding reverses these stages. It first decodes all $b_i$, which determines
which elements participate in each refinement stage and how many raw bits each
element consumes. It then decodes both refinements, unpacks the tail, restores
the implicit leading ones, and applies \textsc{DecodeResidual}.

\paragraph{Entropy tables.}
Modeled symbols use static-table interleaved rANS \citep{duda2013ans}. Counts
are estimated independently for each tensor; only occupied context identifiers
and normalized frequency rows are transmitted. Consequently the codec does
not rely on a model fitted to other checkpoints or external training data. Tables and
headers account for 0.01--0.21\% of compressed output; on the reported BLOOM
delta they occupy 38\,kB alongside approximately 296\,MB of encoded data.

\paragraph{Shared trailing zeros.}
Within each reference-exponent context, ULPack shifts out low-order zero bits
shared by every displacement and records the shift in the header. Checkpoints
trained in bfloat16 but stored in float16, such as Pythia-1B, otherwise pay
for three always-zero bits per weight: the shift lowers Pythia-1B's
100k$\rightarrow$101k ratio from 0.5555 to 0.4319. Where no shift applies,
output is byte-identical to the unshifted format.

\paragraph{Stream layout.}
Each tensor blob is self-contained:
\[
\underbrace{\text{magic, version, count, lane size, section sizes}}_
  {\text{fixed header}}\;
\underbrace{\text{length stage}}_{\text{table and rANS state}}\;
\underbrace{\text{refinement 1}}_{\text{table and rANS state}}\;
\underbrace{\text{refinement 2}}_{\text{table and rANS state}}\;
\underbrace{\text{raw tail}}_{\text{packed bits}}.
\]
Every coded stage records occupied context
ids, normalized frequencies, rANS words, and terminal lane states. The header
also records the exact number of raw-tail bits, making final-byte padding
unambiguous.

\paragraph{Batching.}
Tensors are grouped under a 64\,MiB byte budget and passed through shared rANS
loops to amortize call overhead. Tables, streams, and blobs remain
tensor-local: changing batch boundaries changes neither the compressed bytes
nor decoding independence. Tests assert byte identity across batch sizes,
dtype mixtures, and compiled versus NumPy implementations. This invariant is
required because blobs are content-addressed for deduplication.

\paragraph{GPU implementation.}
ULPack's per-element stages and rANS lanes are independent, so a CUDA
implementation runs them as kernels: residual, exponent, and trailing-zero
extraction; the shift, ZigZag fold, and bit length; refinement contexts;
per-lane rANS encoding with one thread per lane; tail packing; and a decoder
that assigns one block per stream, batched across tensors, and rebuilds the
target tensor directly. Frequency tables are still built on the host by the
CPU code, so GPU blobs are byte-identical to CPU blobs and each side decodes
the other's output. Tests cover all three dtypes at lane-boundary sizes,
bfloat16-grid data with and without the trailing-zero shift, identical
tensors, infinities, NaNs, signed zeros, and subnormals. Beyond the four
checkpoints in Table~\ref{tab:throughput-new}, byte identity also held on
complete Pythia-410M, Pythia-1B, OLMo-2 RLVR, and MultiBERTs pairs
(953 tensors, 6.2\,GB). A single tensor decodes on one streaming
multiprocessor at about 250\,MB/s, so full GPU throughput requires decoding
several tensors at once; encoding is slowest on checkpoints with many small
tensors, where host-side table construction dominates.

\section{Additional experimental details}
\label{app:experiments}

\subsection{Data provenance}

Table~\ref{tab:appendix-data} gives the precise evaluated units. Payload counts
floating-point tensor bytes, not checkpoint container metadata. BLOOM and
Amber are published shards rather than complete models; a shard-delta is also
LMC's unit of evaluation.

\begin{table*}[h]
\centering
\small
\caption{Exact checkpoint pairs and evaluated payloads. Steps are in training
iterations; $t$ ranges over the listed endpoints.}
\label{tab:appendix-data}
\begin{tabular}{lp{0.62\textwidth}r}
\toprule
Sequence & Pairs & Payload \\
\midrule
BLOOM & $t{-}10\text{k}\rightarrow t$ and $t{-}5\text{k}\rightarrow t$ for
$t\in\{20\text{k},30\text{k},\dots,90\text{k}\}$;
90k$\rightarrow$91k, 91k$\rightarrow$92k, 92k$\rightarrow$93k & 4.93\,GB\\
Pythia-70M & all 143 consecutive 1k deltas, 0$\rightarrow$1k through
142k$\rightarrow$143k & 140.9\,MB\\
Pythia-1.4B & $t{-}\Delta\rightarrow t$ for $\Delta\in\{1\text{k},2\text{k},
5\text{k},10\text{k}\}$, $t\in\{20\text{k},40\text{k},\dots,140\text{k}\}$;
100k$\rightarrow$101k; 142k$\rightarrow$143k; warm-up $2^{i}\rightarrow
2^{i+1}$ and 0$\rightarrow$1 (figures only) & 2.83\,GB\\
Pythia 160M--12B & 100k$\rightarrow$101k; also 142k$\rightarrow$143k for 12B;
1B in Appendix~\ref{app:coding} only & up to 23.69\,GB\\
GPT-2 small & 100k$\rightarrow$101k for each of five seeds; for one seed also
100$\rightarrow$200, 1k$\rightarrow$2k, 10k$\rightarrow$11k,
90k$\rightarrow$100k, 95k$\rightarrow$100k, 99k$\rightarrow$100k,
399k$\rightarrow$400k & 248.9\,MB\\
Amber & checkpoints 10$\rightarrow$11, 50$\rightarrow$51, 100$\rightarrow$101 & 13.48\,GB\\
OLMoE & 5k$\rightarrow$10k, 500k$\rightarrow$505k, 1215k$\rightarrow$1220k & 13.84\,GB\\
OLMo & 1k$\rightarrow$2k, 100k$\rightarrow$101k, 390k$\rightarrow$400k,
400k$\rightarrow$401k, 556k$\rightarrow$557k & 27.55\,GB\\
MultiBERTs & 80k$\rightarrow$100k (Table~\ref{tab:main}); all five 20k deltas
in figures & 534.3\,MB\\
OLMo-2 RLVR & 200$\rightarrow$400, 1200$\rightarrow$1400, 2400$\rightarrow$2600 & 2.97\,GB\\
\bottomrule
\end{tabular}
\end{table*}

BLOOM uses \texttt{bigscience/bloom-intermediate}, file
\texttt{pytorch\_model\_00036-of-00072.bin}. The 10k sequence comprises
10k$\rightarrow$20k through 80k$\rightarrow$90k; the matched 5k sequence
comprises 15k$\rightarrow$20k through 85k$\rightarrow$90k; and the late
sequence comprises 90k$\rightarrow$91k through 92k$\rightarrow$93k. Every
protocol uses all 12 bfloat16 tensors from transformer blocks 34 and 35 and
every row in the shard. Amber uses
\texttt{pytorch\_model-00002-of-00002.bin} from revisions
\texttt{ckpt\_100} and \texttt{ckpt\_101}. Pythia and MultiBERTs use complete
models without tensor sampling. OLMo uses \texttt{allenai/OLMo-7B}; GPT-2
small uses the Stanford CRFM runs \texttt{stanford-crfm/alias-gpt2-small-x21},
\texttt{battlestar-gpt2-small-x49}, \texttt{caprica-gpt2-small-x81},
\texttt{darkmatter-gpt2-small-x343}, and \texttt{expanse-gpt2-small-x777};
OLMo-2 RLVR uses \texttt{allenai/OLMo-2-0425-1B-RLVR1}.

\paragraph{Asset licenses.}
BLOOM is under BigScience RAIL v1.0; Amber, Pythia, MultiBERTs, OLMo, OLMoE,
OLMo-2, and the Stanford CRFM GPT-2 small runs are under Apache 2.0; and the
OpenAI GPT-2 and \texttt{gpt2-alpaca-gpt4} checkpoints are under MIT licenses. ZipNN is MIT
licensed and python-zstandard is BSD-3-Clause. The pinned FM-Delta repository
declares no license; we use it only to produce measurements and do not
redistribute its source. Python's wrappers are PSF-2.0 licensed; the underlying
zlib and bzip2 libraries retain their respective zlib and bzip2 licenses.

\subsection{Metric and verification}

For a target checkpoint payload of $N$ bytes and compressed delta of $C$ bytes,
the reported delta ratio is $C/N$. The independently compressed anchor $X_1$
is common to all differential methods and excluded. For baseline size
$C_{\mathrm{base}}$, the reported margin is
$1-C_{\mathrm{ULPack}}/C_{\mathrm{base}}$.

Every method is tested end to end: the compressed residual is decoded, inverted
against its reference, and compared byte-for-byte with the original target
tensor. A failed comparison invalidates the row. Compression ratios are
deterministic; timing runs use the best of two repetitions to reduce incidental
contention.

\subsection{Baseline inputs}

\begin{table}[h]
\centering
\small
\caption{Inputs supplied to each runnable baseline.}
\begin{tabular}{lll}
\toprule
Method & Residual & Preprocessing \\
\midrule
zstd-3, zstd-19 & XOR & bytes grouped by significance\\
gzip-6, bzip2-9 & XOR & bytes grouped by significance\\
ZipNN & XOR & native dtype hint; internal byte grouping\\
FM-Delta & ordered integer & sign/MSB-position coding; low bits raw\\
ULP+zstd-19 ablation & ULP & bytes grouped by significance\\
ULPack & ULP & exponent-conditioned two-part code\\
\bottomrule
\end{tabular}
\end{table}

Generic compressors receive byte-grouped XOR because this is LMC's front end
and is substantially more favorable than the original interleaved byte order.
ZipNN receives plain XOR because its dtype-specific mode performs the grouping
internally. ZipNN mutates bfloat16 and float32 input buffers in place, so each
timed invocation receives a fresh copy prepared outside the timed region.

\paragraph{FM-Delta implementation.}
We use the public \texttt{ningwanyi/FM-Delta} implementation at commit
\texttt{23b2d4297642b8ddfafeaad65824e0751492809e}. For each tensor we call
\texttt{fmd.compress\_param(reference, target)} and include its per-tensor
header in compressed size. The API has no bfloat16 dtype, but its 16-bit path
immediately reinterprets inputs as signed 16-bit integers and performs no
float16 arithmetic; we therefore pass bfloat16 payloads as float16 views of
the same bytes, without numerical conversion. Every decoded tensor is checked
against the original bytes, including synthetic bfloat16 edge-case tests.

\paragraph{Runtime protocol.}
Unless otherwise noted, codecs are evaluated with one thread to avoid
differences in default parallelism: ZipNN otherwise uses multiple cores, while
zstd can parallelize internally. Encode and decode wall-clock times are
measured separately. The original common-subset timing runs used an Apple
M-series host with 10 cores and 34\,GB of memory; full-shard robustness and
ablation runs used a Linux VM with 30 Intel Xeon Platinum 8358 vCPUs and
222\,GiB of memory. All codec measurements were CPU-only, except the ULPack (GPU) row of
Table~\ref{tab:throughput-new}, measured on one NVIDIA A10. We do not compare
throughput across those machines.

\begin{table*}[h]
\centering
\small
\caption{Single-thread performance on the eight full-shard BLOOM 10k-step
deltas. Throughput is aggregate payload divided by aggregate wall time.}
\label{tab:throughput}
\begin{tabular}{lrrr}
\toprule
Method & Compression ratio $\downarrow$ & Encode (MB/s) $\uparrow$ & Decode (MB/s) $\uparrow$\\
\midrule
\textbf{ULPack} & \textbf{0.5705} & 43.2 & 38.9\\
FM-Delta & 0.6130 & 45.9 & 50.2\\
ZipNN & 0.6197 & \textbf{432.4} & 876.7\\
XOR + zstd-19 & 0.6209 & 1.89 & \textbf{952.2}\\
\bottomrule
\end{tabular}
\end{table*}

\begin{table*}[h]
\centering
\scriptsize
\caption{Recorded single-thread encode/decode wall times in seconds on the
Linux VM; dashes denote methods not run in that experiment.}
\label{tab:runtime}
\resizebox{\textwidth}{!}{%
\begin{tabular}{lrrrrr}
\toprule
Sequence & ULPack & ULP+zstd-19 & XOR+zstd-19 & ZipNN & FM-Delta\\
\midrule
BLOOM (10k) & 111.8--116.7 / 125.8--129.6 & --- & 2269.4--2887.7 / 4.1--6.3
& 10.1--12.1 / 3.7--6.5 & 105.4--108.5 / 96.1--101.1\\
BLOOM (5k) & 112.9--121.8 / 128.4--137.8 & --- & 1848.3--2500.5 / 3.4--6.2
& 11.1--12.1 / 4.7--6.5 & 103.7--108.6 / 95.4--101.1\\
BLOOM (1k) & 108.6--112.8 / 127.8--130.4 & 1835.8--2024.9 / 4.0--4.1
& 1823.2--2028.8 / 4.4--4.5 & 11.6--11.7 / 5.7 & 97.3--98.3 / 89.3--91.0\\
Amber-7B & 77.1 / 89.9 & 1187.4 / 3.6 & 1283.4 / 3.8
& 8.3 / 4.3 & 73.2 / 67.6\\
Pythia-70m & 3.61 / 3.52 & --- & --- & --- & 3.01 / 2.77\\
MultiBERTs & 6.72 / 6.28 & --- & --- & --- & 7.52 / 7.20\\
GPT-2 fine-tuning & 3.84 / 3.35 & --- & --- & --- & 4.40 / 4.30\\
\bottomrule
\end{tabular}
}
\end{table*}

\begin{table*}[h]
\centering
\small
\caption{Performance on additional checkpoints (encode / decode, MB/s).
ULPack (GPU) runs on one NVIDIA A10 with tensors resident on the GPU,
excluding one-time compilation; its blobs are byte-identical to ULPack's. All
other rows are single-thread CPU.}
\label{tab:throughput-new}
\begin{tabular}{lrrrr}
\toprule
Method & Pythia-1.4B & Pythia-6.9B & Amber-7B & OLMo-7B\\
\midrule
\textbf{ULPack} & 45.3 / 41.6 & 44.8 / 39.4 & 47.0 / 39.0 & 89.5 / 73.6\\
\textbf{ULPack} (GPU) & 480.5 / 519.4 & 700.4 / 294.6 & 993.6 / 437.3 & 898.8 / 553.6\\
FM-Delta & 47.9 / 52.3 & 45.1 / 50.3 & 49.3 / 51.7 & 71.6 / 77.1\\
ZipNN & 704.1 / 1763.1 & 427.4 / 1408.6 & 423.9 / 813.0 & 424.6 / 1176.4\\
XOR + zstd-19 & 2.8 / 1546.6 & 2.5 / 936.2 & 2.6 / 930.6 & 4.5 / 1335.9\\
ULP + zstd-19 & --- & 2.8 / 855.5 & 2.8 / 982.9 & 4.7 / 1360.4\\
\bottomrule
\end{tabular}
\end{table*}

The reference-context ablation took 6.3--282.5\,s to encode and
4.2--183.6\,s to decode per pair; these uncompiled ablation timings are not
directly comparable to compiled ULPack.

\paragraph{LMC entropy comparison.}
Because LMC does not provide a public implementation, the main-text comparison
uses the blockwise order-0 entropy of its byte-grouped XOR representation over
64\,KiB blocks. For block $B$ with byte counts $n_v$ and $|B|=n$ we compute
\[
  H_0(B)=\sum_{v:n_v>0} n_v\log_2\frac{n}{n_v}.
\]
Summing $H_0(B)$ over blocks and dividing by eight gives the ideal compressed
byte count. This is a favorable lower bound for any memoryless byte coder using
the same block model because it excludes codebook transmission and
integer-code-length overhead.

LMC treats a block specially when all bytes in the 64\,KiB block are identical.
Such blocks account for at most 0.54\% of blocks on any measured delta and
occur in none of the BLOOM or MultiBERTs blocks. This special case therefore
does not materially change the entropy comparison.

\section{Additional compression results}
\label{app:additional-results}

\subsection{Transfer to a fine-tuned GPT-2 model}
\label{app:fine-tuning}

FM-Delta reports \texttt{vicgalle/gpt2-alpaca-gpt4} relative to pretrained
GPT-2 as one of its public atypical-model evaluations \citep{ning2024fm}. We
use the currently published checkpoints pinned at revisions
\texttt{f23966a222c377fe5635293185f83df6504353ff} and
\texttt{607a30d783dfa663caf39e06633721c8d4cfcd7e}, respectively. Following
FM-Delta, we exclude the resized token embedding. We also exclude 12
non-trainable attention-mask buffers whose stored dtypes differ between the
repositories. The remaining 147 shape-compatible float32 parameter tensors
contain 343{,}369{,}728 bytes. FM-Delta is evaluated at revision
\texttt{23b2d4297642b8ddfafeaad65824e0751492809e}.

\begin{table}[h]
\centering
\small
\caption{Compression of compatible parameters in a public base-to-fine-tuned
GPT-2 pair. Ratios are compressed delta size over parameter payload; lower is
better.}
\label{tab:gpt2-fine-tuning}
\begin{tabular}{lrrr}
\toprule
Method & Compressed bytes & Ratio & Relative margin\\
\midrule
\textbf{ULPack} & \textbf{226{,}919{,}912} & \textbf{0.6609} & \textbf{2.9\%}\\
FM-Delta & 233{,}805{,}514 & 0.6809 & ---\\
\bottomrule
\end{tabular}
\end{table}

Both methods reconstruct every tensor bit-exactly. ULPack is smaller on 49 of
147 tensors, and those tensors contain 99.86\% of the evaluated payload.
FM-Delta is smaller on 98 very small tensors, where ULPack's per-tensor tables
and stream metadata dominate. Thus the aggregate gain is broad across
storage-significant parameters but exposes a small-tensor weakness. This
experiment uses FM-Delta's named public model pair, but is not an exact
reproduction of its reported whole-model rate because the paper does not pin
checkpoint revisions and its treatment of excluded payload affects that
denominator.

\subsection{Per-pair results}

Table~\ref{tab:pythia-sizes} compares all Pythia sizes at the same checkpoint
pair. ULPack's margin stays within 6.2--8.2\% across a 170$\times$ range in
parameter count. Table~\ref{tab:warmup-rl} reports the Pythia-1.4B warm-up and
OLMo-2 RLVR pairs, where updates are smallest relative to the weights.

\begin{table}[h]
\centering
\small
\caption{Pythia models at 100k$\rightarrow$101k (lower is better). Pythia-1B
was trained in bfloat16 and stored in float16; ``no shift'' disables the
shared trailing-zero shift (Appendix~\ref{app:coding}).}
\label{tab:pythia-sizes}
\begin{tabular}{lrrrrr}
\toprule
Model & ULPack & FM-Delta & ZipNN & XOR + zstd-19 & Margin\\
\midrule
70M & \textbf{0.6012} & 0.6412 & 0.6791 & 0.6861 & 6.2\%\\
160M & \textbf{0.6122} & 0.6558 & 0.6852 & 0.6893 & 6.6\%\\
410M & \textbf{0.5704} & 0.6080 & 0.6476 & 0.6527 & 6.2\%\\
1B & \textbf{0.4319} & 0.6045 & 0.4704 & 0.4780 & 8.2\%\\
1B (no shift) & 0.5555 & 0.6045 & \textbf{0.4704} & 0.4780 & $-18.1\%$\\
1.4B & \textbf{0.5615} & 0.6010 & 0.6427 & 0.6466 & 6.6\%\\
2.8B & \textbf{0.5561} & 0.5963 & 0.6397 & 0.6444 & 6.7\%\\
6.9B & \textbf{0.5445} & 0.5851 & 0.6332 & 0.6382 & 6.9\%\\
12B & \textbf{0.5472} & 0.5872 & 0.6342 & 0.6399 & 6.8\%\\
\bottomrule
\end{tabular}
\end{table}

\begin{table}[h]
\centering
\small
\caption{Pythia-1.4B warm-up and OLMo-2 RLVR pairs (lower is better). At
step 0$\rightarrow$1 almost no weight changes; ULPack's per-tensor overhead
gives 0.0010 against below 0.0001 for XOR with zstd-19.}
\label{tab:warmup-rl}
\begin{tabular}{lrrrrr}
\toprule
Pair & ULPack & FM-Delta & ZipNN & XOR + zstd-19 & Margin\\
\midrule
\multicolumn{6}{l}{\textit{Pythia-1.4B warm-up}}\\
1$\rightarrow$2 & \textbf{0.0045} & 0.0217 & 0.1336 & 0.0300 & 79.3\%\\
2$\rightarrow$4 & \textbf{0.0346} & 0.1009 & 0.1979 & 0.1233 & 65.7\%\\
4$\rightarrow$8 & \textbf{0.1042} & 0.1868 & 0.2763 & 0.2284 & 44.3\%\\
8$\rightarrow$16 & \textbf{0.2243} & 0.2862 & 0.3807 & 0.3365 & 21.6\%\\
16$\rightarrow$32 & \textbf{0.3138} & 0.3634 & 0.4489 & 0.4146 & 13.7\%\\
32$\rightarrow$64 & \textbf{0.4017} & 0.4439 & 0.5213 & 0.4977 & 9.5\%\\
64$\rightarrow$128 & \textbf{0.4696} & 0.5079 & 0.5722 & 0.5618 & 7.5\%\\
128$\rightarrow$256 & \textbf{0.5449} & 0.5799 & 0.6284 & 0.6239 & 6.0\%\\
256$\rightarrow$512 & \textbf{0.6390} & 0.6724 & 0.7020 & 0.7044 & 5.0\%\\
512$\rightarrow$1k & \textbf{0.7235} & 0.7559 & 0.7650 & 0.7651 & 4.3\%\\
1k$\rightarrow$2k & \textbf{0.7882} & 0.8194 & 0.8201 & 0.8220 & 3.8\%\\
\midrule
\multicolumn{6}{l}{\textit{OLMo-2 1B RLVR}}\\
200$\rightarrow$400 & \textbf{0.0427} & 0.0645 & 0.1617 & 0.0860 & 33.7\%\\
1200$\rightarrow$1400 & \textbf{0.0435} & 0.0611 & 0.1588 & 0.0821 & 28.8\%\\
2400$\rightarrow$2600 & \textbf{0.0477} & 0.0651 & 0.1617 & 0.0871 & 26.6\%\\
\bottomrule
\end{tabular}
\end{table}

\subsection{Additional compressors and throughput}

Table~\ref{tab:additional-codecs} reports calibration codecs on the original
benchmark subsets. These codecs were not rerun on the complete shards and are
not used for the headline margin.

\begin{table}[h]
\centering
\small
\caption{Additional compression ratios; lower is better. BLOOM is the mean of
three 900.4\,MB row samples and Amber is the original 629.2\,MB tensor sample.}
\label{tab:additional-codecs}
\begin{tabular}{lrrrr}
\toprule
Method & BLOOM & Amber & Pythia & MultiBERTs\\
\midrule
XOR + zstd-3 & 0.4333 & 0.5376 & 0.7139 & 0.8328\\
XOR + gzip-6 & 0.4436 & 0.5315 & 0.7102 & 0.8298\\
XOR + bzip2-9 & 0.4588 & 0.5504 & 0.7206 & 0.8315\\
\bottomrule
\end{tabular}
\end{table}

On the same single-threaded runs, ULPack encoded at 75--141\,MB/s and decoded
at 46--132\,MB/s. ZipNN encoded at 773--1093\,MB/s and decoded at
1417--3160\,MB/s. These timings are of the original Python implementation;
current CPU and GPU throughput is reported in Tables~\ref{tab:throughput}
and~\ref{tab:throughput-new}.

\section{Bit-domain analysis}
\label{app:bit-domain}

\subsection{Reference-exponent ablation}

Table~\ref{tab:context-ablation} removes $e_r$ from all three probability
models while retaining the ULP residual, symbol decomposition, rANS coder,
per-tensor tables, raw-tail layout, and batching. The ablated models are
$P(b)$, $P(q_1\mid b)$, and $P(q_2\mid b,q_1)$. Thus the measured difference
isolates reference-exponent conditioning rather than the ordered-integer
representation or coder. Without $e_r$, ULPack is within 1\% of FM-Delta on
every row, so its advantage over FM-Delta comes from the reference exponent.

\begin{table}[h]
\centering
\small
\caption{Controlled ablation of reference-exponent conditioning. BLOOM is the
mean of the three complete-shard 1k deltas. ``Reduction'' is the decrease in
output from restoring $e_r$; every row reconstructs bit-exactly.}
\label{tab:context-ablation}
\begin{tabular}{lrrr}
\toprule
Sequence & Without $e_r$ & ULPack & Reduction\\
\midrule
BLOOM (1k) & 0.3662 & \textbf{0.3255} & 11.1\%\\
Amber (shard) & 0.4616 & \textbf{0.4167} & 9.7\%\\
Amber (ckpt 100$\rightarrow$101) & 0.4659 & \textbf{0.4201} & 9.8\%\\
Pythia-70m & 0.6402 & \textbf{0.6012} & 6.1\%\\
Pythia-6.9B & 0.5842 & \textbf{0.5445} & 6.8\%\\
Pythia-12B & 0.5866 & \textbf{0.5472} & 6.7\%\\
OLMo-7B & 0.7237 & \textbf{0.7063} & 2.4\%\\
MultiBERTs & 0.8051 & \textbf{0.7920} & 1.6\%\\
\bottomrule
\end{tabular}
\end{table}

\subsection{Residual bit positions}

\begin{figure}[h]
\centering
\includegraphics[width=\linewidth]{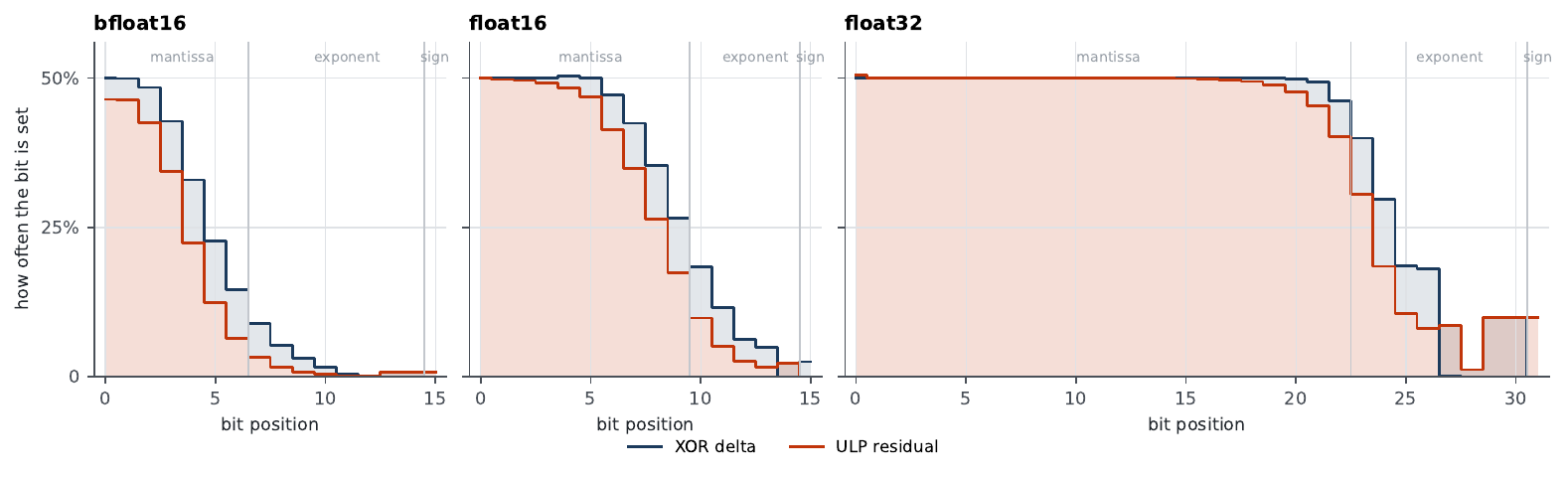}
\caption{Per-position set frequency in XOR and ULP representations. The area
between curves shows where rank-space differencing shortens the representation
before context modeling.}
\label{fig:bits-appendix}
\end{figure}

Figure~\ref{fig:bits-appendix} measures how often each bit is set in the XOR
delta and ULP residual. Moving to rank space generally shifts mass toward less
significant positions. The exception is float32: frequent zero crossings move
mass into the top of the rank-space word, so independent bit coding is slightly
worse in ULP space even though ULPack remains smaller after conditioning.

Summing independent binary entropies gives 7.435, 11.751, and 26.701
bits/weight for XOR on bfloat16, float16, and float32; ULP gives 6.220, 10.617,
and 27.348. ULPack's achieved rates are 5.270, 9.619, and 25.344 bits/weight.
The float32 result is important: changing domain alone does not explain the
gain there; exponent conditioning does.

The LMC-model bound gives a complementary calibration on the original
common-subset analysis:
\begin{table}[h]
\centering
\small
\caption{Ideal order-0 byte coding of byte-grouped XOR compared with ULPack.
BLOOM uses three 900.4\,MB row subsets and Amber the original 629.2\,MB tensor
subset; these calibration units therefore do not match the complete-shard
headline rows in Table~\ref{tab:main}.}
\begin{tabular}{lrrr}
\toprule
Sequence & XOR byte bound & ULPack & Below bound\\
\midrule
BLOOM-176B & 0.4091 & 0.3294 & 19.5\%\\
Amber & 0.4941 & 0.4185 & 15.3\%\\
Pythia-70m & 0.6679 & 0.6012 & 10.0\%\\
MultiBERTs & 0.8084 & 0.7920 & 2.0\%\\
\bottomrule
\end{tabular}
\end{table}

\section{Trajectory details}
\label{app:trajectory-details}

\subsection{BLOOM frequency and late-stage results}

\paragraph{Matched 5k/10k pairs.}
Table~\ref{tab:bloom-frequency-detail} reports the full-shard measurements
underlying the main-text checkpoint-spacing summary. Within each cell, the 5k
result precedes the matched 10k result.

\begin{table*}[h]
\centering
\small
\caption{Compression ratios at matched BLOOM training stages, shown as
5k / 10k.}
\label{tab:bloom-frequency-detail}
\begin{tabular}{lrrrr}
\toprule
Target & ULPack & FM-Delta & XOR+zstd-19 & ZipNN\\
\midrule
20k & 0.6156 / 0.6682 & 0.6592 / 0.7168 & 0.6420 / 0.6761 & 0.6589 / 0.6912\\
30k & 0.5913 / 0.6348 & 0.6335 / 0.6798 & 0.6375 / 0.6569 & 0.6279 / 0.6778\\
40k & 0.5702 / 0.6114 & 0.6115 / 0.6545 & 0.6282 / 0.6473 & 0.6120 / 0.6517\\
50k & 0.5481 / 0.5890 & 0.5888 / 0.6307 & 0.6128 / 0.6386 & 0.5978 / 0.6248\\
60k & 0.5229 / 0.5642 & 0.5633 / 0.6050 & 0.5945 / 0.6224 & 0.5830 / 0.6077\\
70k & 0.4928 / 0.5352 & 0.5332 / 0.5754 & 0.5712 / 0.6041 & 0.5648 / 0.5900\\
80k & 0.4568 / 0.5006 & 0.4973 / 0.5407 & 0.5407 / 0.5783 & 0.5432 / 0.5694\\
90k & 0.4157 / 0.4605 & 0.4567 / 0.5007 & 0.5029 / 0.5437 & 0.5162 / 0.5453\\
\bottomrule
\end{tabular}
\end{table*}

\paragraph{Late 1k sequence.}
\begin{table*}[h]
\centering
\small
\caption{Compression ratios on complete BLOOM shard 36 at 1k-step intervals,
with ULPack's relative reduction against each baseline in parentheses.}
\label{tab:bloom-1k}
\begin{tabular}{lrrrr}
\toprule
Pair & ULPack & FM-Delta & XOR+zstd-19 & ZipNN\\
\midrule
90k--91k & \textbf{0.3294} & 0.3716 (11.4\%) & 0.4170 (21.0\%) & 0.4515 (27.0\%)\\
91k--92k & \textbf{0.3255} & 0.3677 (11.5\%) & 0.4135 (21.3\%) & 0.4482 (27.4\%)\\
92k--93k & \textbf{0.3216} & 0.3639 (11.6\%) & 0.4097 (21.5\%) & 0.4451 (27.7\%)\\
\midrule
Mean & \textbf{0.3255} & 0.3677 (11.5\%) & 0.4134 (21.3\%) & 0.4483 (27.4\%)\\
\bottomrule
\end{tabular}
\end{table*}

\begin{figure}[h]
\centering
\includegraphics[width=0.7\linewidth]{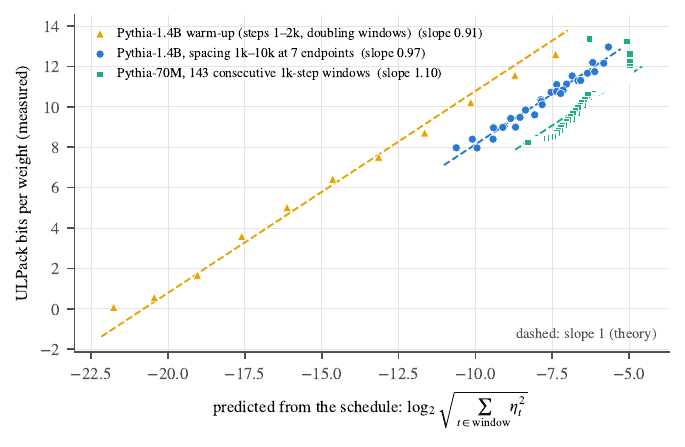}
\caption{ULPack bits per weight against the learning rate accumulated over
the window between checkpoints.}
\label{fig:spacing}
\end{figure}

\paragraph{Pythia spacing.}
Figure~\ref{fig:spacing} reports the Pythia-1.4B spacing pairs and the
log-spaced warm-up checkpoints against the learning rate accumulated over each
window. Within each set, bits per weight rise with slope 0.91--0.97 in
$\log_2\sqrt{\sum_t\eta_t^2}$.

\subsection{Where the encoded bits go}

Because ULPack emits the raw tail verbatim, the modeled share bounds what
improved models of its current coded stages can save. The monotone relationship in
Table~\ref{tab:conditioning} explains the format dependence more precisely
than significand width alone: convergence also determines how many high
residual bits remain structured.

\subsection{Evolution within and across tensors}
\begin{figure*}[t]
\centering
\includegraphics[width=\textwidth]{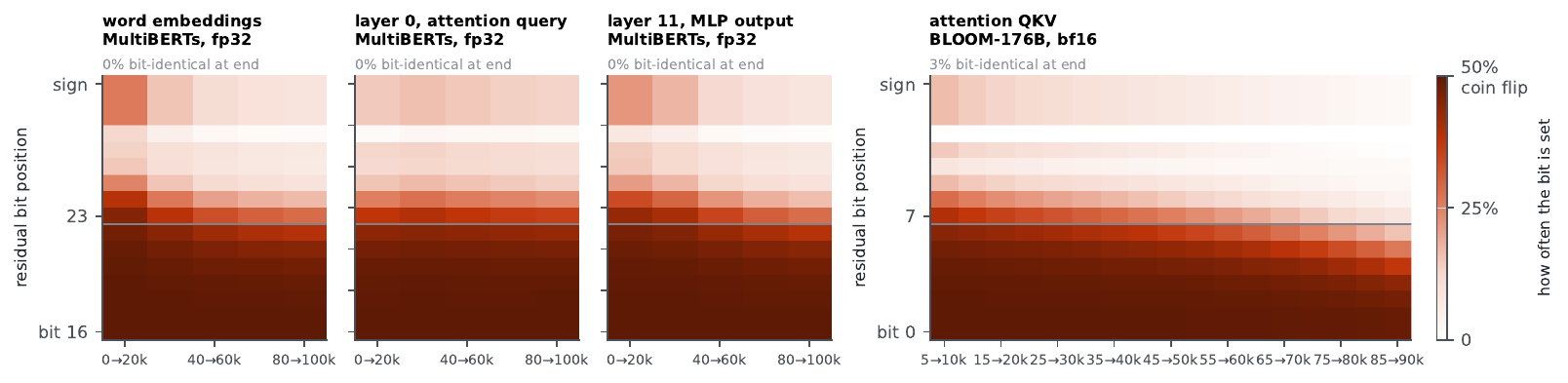}
\caption{Residual-bit activity through training. BLOOM tracks the leading
18\% of rows of a representative attention-QKV tensor; high-order activity
recedes while low-order residual bits remain near the 50\% coin-flip limit.
MultiBERTs shows that convergence differs across parameter roles.}
\label{fig:evolution}
\end{figure*}

The BLOOM panel of Figure~\ref{fig:evolution} follows the leading 18\% of rows
of one attention QKV tensor over 17 uniformly spaced 5k-step deltas. Its
zero-crossing rate falls from 16.6\% to 2.1\%, while its lowest-order residual
bit remains near 50\%.
MultiBERTs shows that convergence is not uniform across parameter roles:
embeddings and a late MLP output settle substantially, while an early
attention-query tensor changes little.

\subsection{Previous-step information}

Let $b_t$ denote current residual bit length. Table~\ref{tab:temporal-mi}
separates information in the previous residual from information already
provided by the reference exponent.

\begin{table}[h]
\centering
\small
\caption{Mutual information with current residual length, in bits per weight.}
\label{tab:temporal-mi}
\begin{tabular}{lrrrr}
\toprule
Sequence & $H(b_t)$ & $I(b_t;b_{t-1})$ & $I(b_t;e_t)$
  & $I(b_t;b_{t-1}\mid e_t)$\\
\midrule
BLOOM-176B & 3.0598 & 0.2979 & 0.6240 & 0.0006\\
MultiBERTs & 3.1302 & 0.1594 & 0.4389 & 0.0020\\
\bottomrule
\end{tabular}
\end{table}

Previous-step scale is informative alone, but contributes only 0.02\% and
0.06\% of $H(b_t)$ after exponent conditioning. Thus the phase portrait's
magnitude dependence is real but almost entirely redundant with ULPack's
existing free context.

\section{Alternatives that did not help}
\label{app:alternatives}

\begin{figure*}[t]
\centering
\includegraphics[width=0.9\textwidth]{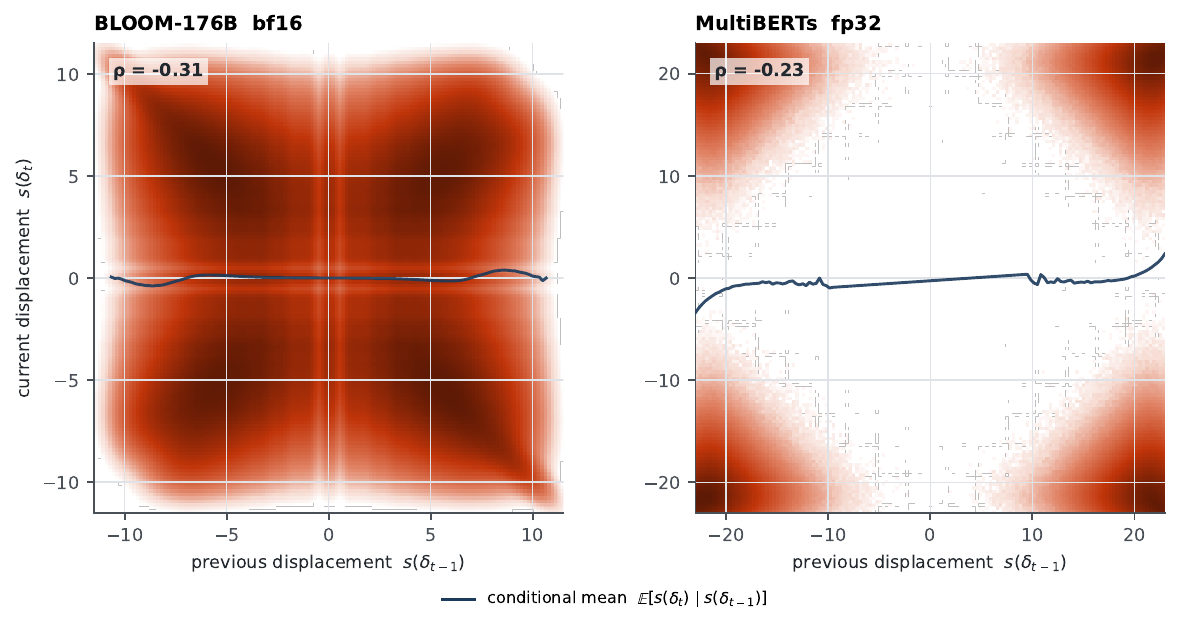}
\caption{Phase portraits of consecutive rank-space displacements. Color shows
joint density and the blue curve shows the conditional mean after a signed-log
display transform. The reported correlations are computed on the raw signed
displacements; their weak negative values indicate little reusable momentum
beyond the immediate reference.}
\label{fig:phase}
\end{figure*}

\paragraph{Previous residual as context.}
Conditioning directly on $b_{t-1}$ would make a blob depend on its grandparent
checkpoint and weaken content-addressed deduplication. Table~\ref{tab:temporal-mi}
shows that this cost buys at most 0.002 bits per weight beyond the exponent.

\paragraph{Bidirectional and momentum prediction.}
For three consecutive checkpoints $(a,b,c)$, we compare the shipped chain
$C(b\mid a)+C(c\mid b)$ with an anchor-plus-bridge arrangement
$C(c\mid a)+C(b\mid a,c)$. The middle frame becomes cheaper, but the doubled
anchor gap costs more:
\begin{table}[h]
\centering
\small
\caption{Idealized trajectory-prediction costs in bits per weight.}
\begin{tabular}{lrrrr}
\toprule
Sequence & $\rho$ & Shipped chain & Anchor+bridge & Change\\
\midrule
BLOOM QKV, 5k gaps & $-0.31$ & 17.273 & 17.411 & $+0.80\%$\\
MultiBERTs, 20k gaps & $-0.23$ & 51.479 & 51.824 & $+0.67\%$\\
\bottomrule
\end{tabular}
\end{table}
All 20 measured triples lose. Under a Gaussian approximation, BLOOM's
correlation contains
$-\tfrac12\log_2(1-\rho^2)=0.073$ bits per weight, only 0.42\% of its
bfloat16 chain. A one-sided momentum predictor is worse still, increasing
forward cost by 2.40\%.

\paragraph{Additional refinement levels.}
ULPack rANS-codes $b$, $q_1$, and $q_2$, then writes lower bits verbatim. We
tested refinement depths through six by measuring the conditional entropy of
each bit under the context available to that stage. Table~\ref{tab:refinement-depth}
reports the resulting upper bound on savings relative to leaving the bit raw.

\begin{table}[h]
\centering
\small
\caption{Conditional entropy and optimistic saving at each refinement level.
Savings are percentages of uncompressed payload and exclude probability-table
costs and the additional pass required by each level. The bfloat16 measurement
uses a 157.3\,MB Amber sample; float32 uses the complete 534.3\,MB final
MultiBERTs delta. Levels 1 and 2 are used by ULPack.}
\label{tab:refinement-depth}
\begin{tabular}{crrrr}
\toprule
& \multicolumn{2}{c}{bfloat16} & \multicolumn{2}{c}{float32}\\
\cmidrule(lr){2-3}\cmidrule(lr){4-5}
Level & $H(q_\ell\mid\mathrm{ctx})$ & Saving & $H(q_\ell\mid\mathrm{ctx})$ & Saving\\
\midrule
1 & 0.9170 & 0.47\% & 0.8543 & 0.46\%\\
2 & 0.9468 & 0.28\% & 0.8864 & 0.36\%\\
3 & 0.9515 & 0.22\% & 0.9366 & 0.20\%\\
4 & 0.9548 & 0.16\% & 0.9095 & 0.28\%\\
5 & 0.9797 & 0.05\% & 0.9497 & 0.16\%\\
6 & 0.9719 & 0.05\% & 0.9727 & 0.09\%\\
\bottomrule
\end{tabular}
\end{table}

Levels 3--6 together save only 0.48\% of bfloat16 payload and 0.73\% of
float32 payload. Relative to compressed size, these are approximately 1.1\%
and 0.9\%, respectively, even before charging for four additional context and
rANS passes. Their bits are already close to uniform, so the small ratio gain
does not justify the added compute.

\paragraph{Magnitude-space handling of zero crossings.}
Rank distance is large when a parameter crosses zero. A reversible alternative
that differences magnitudes and codes the sign transition separately reduces
estimated size by only 0.13--1.41\% across datasets, far below the apparent
cost of crossing weights. Their associated low-order residual bits carry
nearly full entropy, so most of that cost is information rather than coding
slack.

\newpage

\end{document}